\documentclass{article} 
\usepackage{iclr2026_conference,times}

\usepackage{amsmath,amsfonts,bm}

\def\eqref#1{equation~\ref{#1}}

\def\1{\bm{1}}

\DeclareMathAlphabet{\mathsfit}{\encodingdefault}{\sfdefault}{m}{sl}
\SetMathAlphabet{\mathsfit}{bold}{\encodingdefault}{\sfdefault}{bx}{n}

\usepackage{hyperref}
\usepackage{url}
\usepackage{algorithm}
\usepackage{algpseudocode}
\usepackage{amssymb}
\usepackage{booktabs}
\usepackage{enumitem}
\usepackage{listings}
\usepackage{xcolor}        
\usepackage{graphicx}
\usepackage{colortbl}
\usepackage{pifont}
\usepackage[most]{tcolorbox}
\usepackage{amsmath}

\lstdefinestyle{custompython}{
    language=Python,
    backgroundcolor=\color{gray!15},  
    commentstyle=\color{gray!50!green}, 
    keywordstyle=\color{purple!80!black},  
    numberstyle=\tiny\color{gray!50!black},  
    stringstyle=\color{green!70!black},      
    basicstyle=\ttfamily\footnotesize\color{gray!90!black},  
    breaklines=true,
    frame=single,
    numbers=left,
    showstringspaces=false,
    captionpos=b
}

\title{MemOrb: A Plug-and-Play Verbal-Reinforcement Memory Layer for E-Commerce Customer Service}

\author{
Yizhe Huang\\
\textsuperscript{\rm1}Xiaoduo AI, \textsuperscript{\rm2}Fudan University \\
\texttt{24210720111@m.fudan.edu.cn} \\
\And
Yang Liu \\
\textsuperscript{\rm1}Xiaoduo AI \\
\texttt{liuyangfoam@xiaoduotech.com} \\
\And
Ruiyu Zhao \\
\textsuperscript{\rm1}Xiaoduo AI, \textsuperscript{\rm3}East China University of Science and Technology \\
\texttt{zhaoruiyu96@gmail.com} \\
\And
Xiaolong Zhong \\
\textsuperscript{\rm1}Xiaoduo AI, \textsuperscript{\rm4}Shanghai Jiao Tong University \\
\texttt{zhongxiaolong@xiaoduotech.com} \\
\And
Xingming Yue \\
\textsuperscript{\rm1}Xiaoduo AI \\
\texttt{yuexingming@xiaoduotech.com} \\
\And
Ling Jiang \\
\textsuperscript{\rm1}Xiaoduo AI \\
\texttt{jiangling@xiaoduotech.com}
}

\makeatletter
\def\@oddhead{}
\def\@evenhead{}
\makeatother

\iclrfinalcopy 
\begin{document}

\maketitle

\begin{abstract}
Large Language Model-based agents(LLM-based agents) are increasingly deployed in customer service, yet they often forget across sessions, repeat errors, and lack mechanisms for continual self-improvement. This makes them unreliable in dynamic settings where stability and consistency are critical. To better evaluate these properties, we emphasize two indicators: \textit{task success rate} as a measure of overall effectiveness, and \textit{consistency metrics} such as Pass$^k$ to capture reliability across multiple trials. To address the limitations of existing approaches, we propose \textbf{MemOrb}, a lightweight and plug-and-play verbal reinforcement memory layer that distills multi-turn interactions into compact strategy reflections. These reflections are stored in a shared memory bank and retrieved to guide decision-making, without requiring any fine-tuning. Experiments show that MemOrb significantly improves both success rate and stability, achieving up to a 63 percentage-point gain in multi-turn success rate and delivering more consistent performance across repeated trials. Our results demonstrate that structured reflection is a powerful mechanism for enhancing long-term reliability of frozen LLM agents in customer service scenarios.
\end{abstract}
\section{Introduction}
\label{sec:intro}
Large Language Model-based agents (LLM-based agents) are increasingly adopted in large-scale customer service systems, where they act as interactive assistants for diverse users~\citep{brown2020language}. Despite their rapid deployment, these agents face persistent challenges: they often lose critical information across sessions, repeat errors without systematic correction, and struggle to adapt to rapidly changing product catalogs. Such limitations undermine their reliability in dynamic environments such as e-commerce.

Existing memory solutions typically rely on short-term caching or user-specific profiles~\citep{chhikara2025mem0buildingproductionreadyai, zhong2023memorybankenhancinglargelanguage}. While these approaches can temporarily capture context or recall user preferences, they fail in e-commerce scenarios where fewer than 5\% of queries recur and thousands of new products appear daily. Consequently, purely per-user or short-horizon memories are insufficient for robust long-term improvement.

Recent advances in fine-tuning and reinforcement learning have shown progress in domains such as mathematics, finance, and healthcare~\citep{yao2023reactsynergizingreasoningacting, packer2024memgptllmsoperatingsystems}. However, applying these methods to customer service is challenging: (1) the domain involves heterogeneous buyer profiles, frequent return/exchange requests, and logistics queries requiring complex tool invocation; and (2) the cost of continual fine-tuning or reinforcement learning is computationally prohibitive for production systems that demand rapid adaptation.

To address these challenges, we introduce \textbf{MemOrb}, a plug-and-play memory layer designed to transform frozen LLM-based agents into continuously improving assistants. MemOrb builds on the \textbf{reflexion paradigm}~\citep{shinn2023reflexionlanguageagentsverbal}, which generates structured reflections from completed tasks and reuses them for future decision-making. Unlike prior frameworks such as Mem0~\citep{chhikara2025mem0buildingproductionreadyai}, A-Mem~\citep{xu2025amemagenticmemoryllm}, or skill-code repositories like Voyager~\citep{wang2023voyageropenendedembodiedagent} and Optimus-1~\citep{li2024optimus1hybridmultimodalmemory}, MemOrb captures policy-level reflections that can be shared across users without schema constraints. 

After each dialogue, MemOrb condenses the interaction—including tool usage—into a compact \emph{strategy reflection}. This reflection is encapsulated into a memory unit called an \textbf{Orb}, stored in a lightweight SQLite database and indexed in ChromaDB for efficient semantic retrieval. At inference time, the agent reformulates the current user query by incorporating dialogue context, retrieves the most relevant reflections, and integrates them into its prompt. This requires no gradient updates and enables continual self-improvement. In this way, MemOrb reduces reliance on handcrafted prompts and provides a systematic mechanism for accumulating and transferring knowledge across users.

\begin{figure*}[t]
  \centering
  \includegraphics[width=\textwidth]{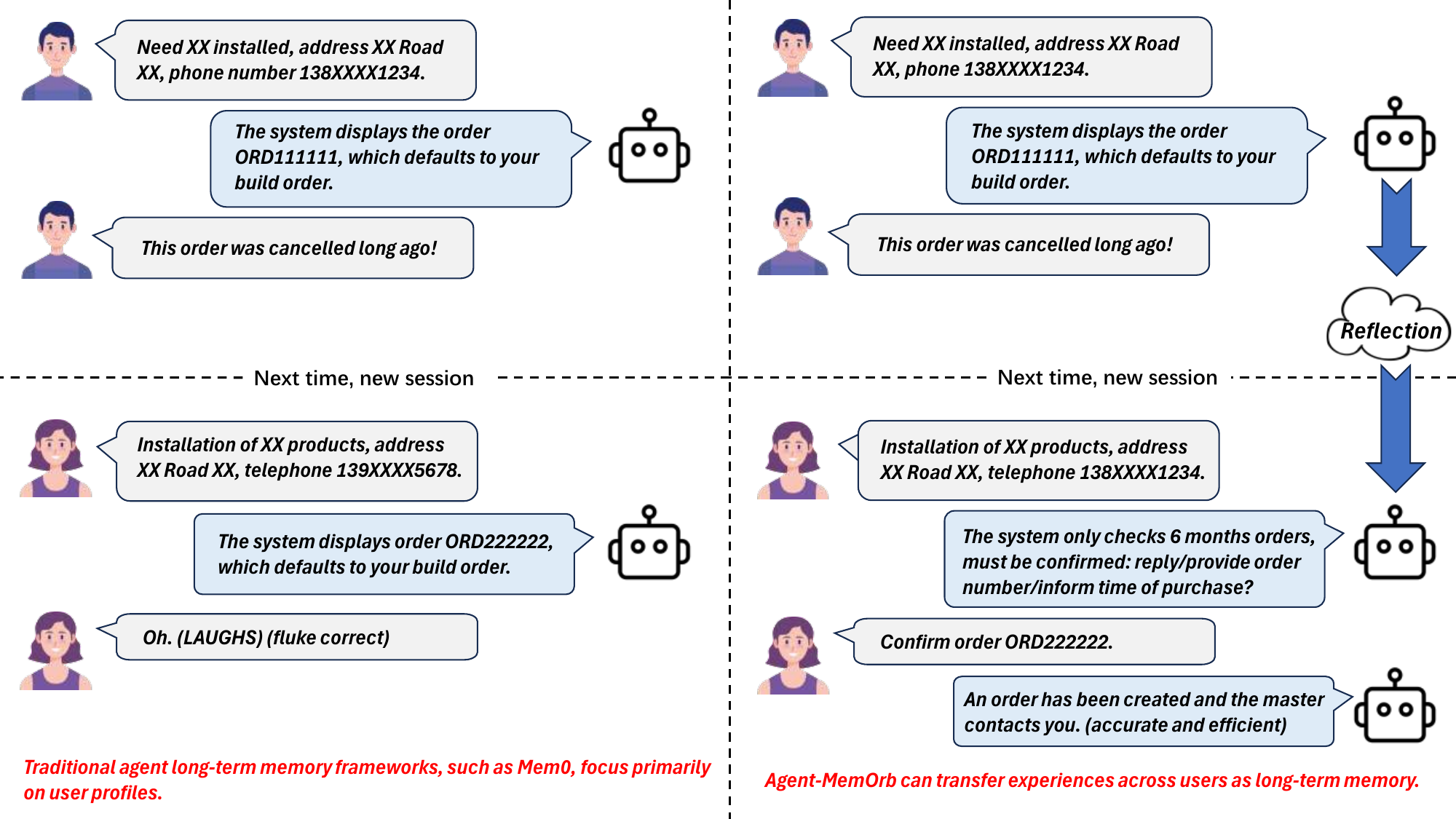}
  \caption{Sample customer service dialogues without memory and with MemOrb, we can see that the intelligent body with MemOrb performs better}
\end{figure*}

To summarize, our contributions are:

\begin{itemize}
    \item We propose \textbf{MemOrb}, a lightweight and schema-free verbal reinforcement memory layer for \textbf{LLM-based agents}, enabling continual improvement without parameter updates by distilling interactions into compact strategy units (Orbs) for efficient cross-user transfer and error reduction.
    
    \item We develop a retrieval and rewriting pipeline that integrates Orbs into prompts at inference, improving decision-making while remaining computationally efficient.
    
    \item We extend \textbf{ECom-Bench} with 77 clothing-domain tasks, creating a total of 130 realistic multi-turn customer service tasks, and conduct extensive experiments demonstrating substantial gains in task success rate and consistency (e.g., up to 63 percentage points improvement).
    
    \item We release an open-source implementation including database schema, retrieval pipeline, and integration toolkit, supporting practical deployment of self-improving LLM-based agents.
\end{itemize}

\section{Related work}

\label{sec:related}
Early Large Language Model (LLM)-based agents primarily relied on short-term sliding windows for context retention \citep{brown2020language,sumers2024cognitivearchitectureslanguageagents}. While effective in short-term dialogues, this approach suffers from context loss in multi-turn interactions, limiting the model's long-term performance. To address this limitation, recent memory architectures can be broadly grouped into four categories.
\paragraph{User-Centric Long-Term Memory}
\textbf{User-centric long-term memory} approaches, such as Mem0 \citep{chhikara2025mem0buildingproductionreadyai} and LangMem, maintain per-user profiles in graph databases, which help recall user preferences and past interactions \citep{park2023generativeagentsinteractivesimulacra}. These methods are well-suited for personalized systems but tend to degrade when queries drift or when facing dynamic environments. Furthermore, they struggle when dealing with large-scale systems where user preferences and interactions frequently change, leading to outdated information and suboptimal performance.
\paragraph{Episodic Retrieval-Augmented Generation}
To overcome the limitations of user-centric memories, \textbf{episodic retrieval-augmented generation} methods, such as MemoryBank \citep{zhong2023memorybankenhancinglargelanguage} and ReadAgent \citep{lee2024humaninspiredreadingagentgist}, store raw dialogue chunks for vector retrieval, avoiding forgetting \citep{guo2024empoweringworkingmemorylarge}. However, these methods often result in context bloat and high token costs as they store large amounts of dialogue history for future retrieval. While they avoid forgetting and improve context retention, they become computationally expensive and inefficient in large-scale systems.
\paragraph{Programmatic Memory Layers}
\textbf{Programmatic memory layers}, including MemGPT \citep{packer2024memgptllmsoperatingsystems} and A-Mem \citep{xu2025amemagenticmemoryllm}, allow explicit or dynamic read/write operations on memory, providing flexibility and improving the adaptability of LLMs in diverse tasks \citep{yan2025memoryr1enhancinglargelanguage}. However, these methods are typically tied to predefined schemas, which can limit their generalization and flexibility when faced with highly variable tasks. Moreover, the integration of such memory systems often requires extensive computational resources and is prone to complexity in real-time systems.
\paragraph{Skill-Code Repositories}
Finally, \textbf{skill-code repositories}, such as Voyager \citep{wang2023voyageropenendedembodiedagent} and Optimus-1 \citep{li2024optimus1hybridmultimodalmemory}, capture executable code for high-level task planning \citep{zhang2024surveymemorymechanismlarge}. These methods focus on storing reusable code that can be invoked to solve a broad range of tasks. However, they lack fine-grained dialogue capabilities and struggle to maintain long-term consistency in conversational agents that require nuanced multi-turn dialogue.
\paragraph{MemOrb: A Reflection-Centered Approach}
In contrast to the aforementioned approaches, \textbf{MemOrb} takes a reflection-centered approach. Instead of indexing raw history or building per-user graphs, it distills \emph{policy reflections} from all conversations into a single, shared memory \citep{shinn2023reflexionlanguageagentsverbal,Renze_2024,bo2024reflective}. This design enables schema-free, cross-user continual learning and reduces reliance on handcrafted prompts or costly fine-tuning. MemOrb offers a lightweight, plug-and-play solution that improves the performance of LLM-based agents without the need for frequent model updates or large-scale retraining, making it particularly suitable for real-world applications where efficiency and scalability are critical.
In summary, while various memory architectures address specific challenges in long-term context retention, each has its trade-offs in terms of scalability, efficiency, and complexity \citep{wu2025humanmemoryaimemory}. MemOrb's reflection-centered design provides a lightweight, flexible, and scalable alternative that overcomes many of the limitations of prior approaches.
\begin{table}[t]
\centering
\small
\caption{Comparison of memory architectures. MemOrb is the only system that combines schema-free policy reflections with continual cross-user learning.}
\begin{tabular}{lccccc}
\toprule
\textbf{System} & \textbf{Storage} & \textbf{Granularity} & \textbf{Re-Write} & \textbf{Cross-User} & \textbf{Schema} \\
\midrule
Mem0 & Graph DB & User profile & $\times$ & $\times$ & Predefined \\
LangMem & KV store & User profile & $\times$ & $\times$ & Predefined \\
MemoryBank & Vector DB & Raw dialogue & $\times$ & $\times$ & None \\
MemGPT & Key-value & Raw dialogue & $\times$ & $\times$ & Predefined \\
A-Mem & Vector DB & Structured events & $\times$ & $\times$ & Dynamic \\
\rowcolor{blue!10}
MemOrb & SQLite+ChromaDB & \textbf{Policy reflection} & $\checkmark$ & $\checkmark$ & Schema-free \\
\bottomrule
\end{tabular}
\label{tab:compare}
\end{table}
\section{Method}
\label{sec:method}

We describe the concrete implementation of \textbf{MemOrb} that converts raw dialogues into compact, queryable memories and injects them back into an agent at inference time.  The system keeps two stores: (1) a \emph{metadata store} backed by SQLite for reliable, low-overhead persistence, and (2) a \emph{vector store} backed by \textsc{ChromaDB} for fast semantic retrieval.  

\begin{figure*}[t]
  \centering
  \includegraphics[width=\textwidth]{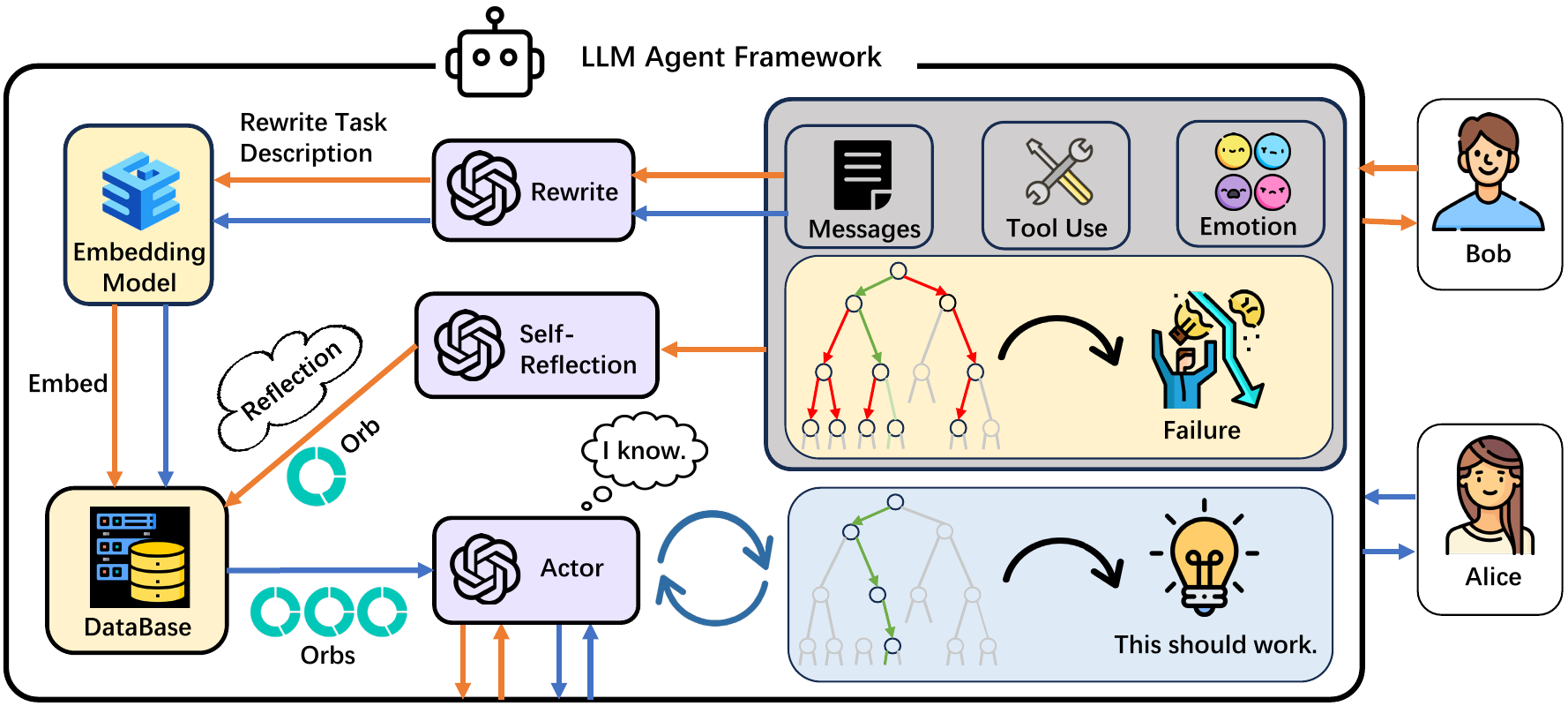}
  \caption{MemOrb data-flow overview: After each episode, trajectories are distilled into policy reflections and stored in a shared memory bank.  During the next turn, the agent rewrites the user query, retrieves relevant memories, and augments its prompt for continual self-evolution.}
  \label{fig:arch}
\end{figure*}

\subsection{MemOrb FrameWork}
Figure~\ref{fig:arch} illustrates the data flow of MemOrb. In the MemOrb framework, we use a modular formula, which includes three models with divided responsibilities: an \textit{Actor} model responsible for generating text and actions, a \textit{Rewrite} model responsible for rewriting the context of customer service user conversations and generating task descriptions, and a \textit{Self-Reflection} model used to reflect on the trajectory and generate Orbs.
\paragraph{Actor.}
The Actor is a Large Language Model(LLM) that generates specific actions and text based on a given prompt and observations of environmental changes. Following the \textbf{ReAct} framework~\citep{yao2023reactsynergizingreasoningacting}, it can take an action $a_t$ according to the policy $\pi_\theta$ at time step $t$, or produce a piece of text and subsequently observe the outcome $o_t$ from the environment. In \textbf{MemOrb}, The Actor model can also retrieve the top-k similar Orbs from the database: these may belong to the same user or different users, in order to construct an enhanced prompt that will standardize the subsequent actions of the Actor.

\paragraph{Rewrite.}
The Rewrite module is used to rewrite users' queries and context descriptions. These rewritten contents are embedded into high-dimensional vectors through an embedding model, and together with the Orbs generated by the Self-Reflection module, they are input into the DataBase for Actor retrieval.

\paragraph{Self-Reflextion.}
Self-Reflection is instantiated as a Large Language Model(LLM) and plays a crucial role in the \textbf{MemOrb} framework. Conventional self-reflection modules are typically used to generate simple self-assessments, but in \textbf{MemOrb} we extend this by incorporating an evaluation module and an Orb generation module. After the completion of a task, the system reflects on both the outcome and the trajectory, where the trajectory includes dialogue context, tool calls, and user emotions. The generated reflections summarize possible reasons for task failure as well as plans for future attempts, and are subsequently stored in the database. For example, in an e-commerce customer service dialogue, the Actor may execute an incorrect action $a_t$, such as providing wrong parameters, invoking the wrong tool, or producing an erroneous output, which then leads to subsequent actions $a_{t+1}, a_{t+2}$. The Self-Reflection module generates reflections on these errors, so that when facing the same or similar tasks in the future, the Actor is more likely to take the improved action $a_t'$, and consequently generate $a_{t+1}'$ and $a_{t+2}'$.

\subsection{Memory Unit: Orb}
\label{sec:orb}

We compress each conversation episode into a lightweight, query-ready structure called an \textbf{Orb}.  
Formally, an Orb is the 6-tuple  
\[
O = \langle \text{id},\ \text{obs},\ \text{emotion},\ \text{outcome},\ \text{context},\ \text{timestamp} \rangle,
\]  
where id is a SHA-256 digest of the concatenation of the remaining fields; obs is the user utterance or system prompt; emotion is a categorical label (e.g., frustrated); outcome is the distilled policy reflection generated by the frozen LLM; context is an optional JSON blob (budget, product SKU, \emph{etc.}); timestamp is the creation time.

\begin{algorithm}[t]
\caption{Orb Generation: Policy-Reflection Distillation}
\label{alg:orb}
\begin{algorithmic}[1]
\Require{Episode trajectory $\tau = \{(u_t, a_t, r_t)\}_{t=1}^T$,\\ frozen LLM $\mathcal{M}$}
\Ensure{Orb $O = \langle \text{id},\ \text{obs},\ \text{emotion},\ \text{outcome},\ \text{context},\ \text{timestamp} \rangle$}
\State $\text{obs} \gets \text{concatenate}(u_1, \dots, u_T)$
\State $\text{emotion} \gets \text{EmotionTagger}(u_T)$
\State $\text{outcome} \gets \text{PolicyReflection}(\mathcal{M}, \tau)$ \Comment{LLM-generated}
\State $\text{context} \gets \text{JSON}(r_T, \text{metadata})$ \Comment{SKU, budget, \dots}
\State $\text{timestamp} \gets \text{now()}$
\State $\text{id} \gets \text{SHA256}(\text{obs} \parallel \text{emotion} \parallel \text{outcome})$
\State \Return $O$
\end{algorithmic}
\end{algorithm}

Algorithm~\ref{alg:orb} depicts the full process.  
Each field is constrained by a lightweight SQLAlchemy model, ensuring consistency across SQLite and ChromaDB.

\subsection{Memory System}
\label{subsec:memory-system}

The memory system of \textbf{MemOrb} consists of three components: a storage layer, a reflection pipeline, and a retrieval pipeline. Together, they enable the agent to store, distill, and reuse experiences efficiently.

\paragraph{Storage Layer.} 
The \textbf{Metadata Store} wraps SQLAlchemy and provides two basic operations: saving an Orb (which upserts a row into the orbs table) and fetching an Orb by its primary key.  
The \textbf{Vector Store} initialises a \textsc{ChromaDB} persistent client at path \texttt{./chroma\_db}. Each Orb is serialised into a document
\begin{equation}
\text{doc} = \text{obs} \oplus \text{emotion} \oplus \text{outcome} \oplus \text{str(context)} ,
\end{equation}
where \textit{obs} denotes the user utterance or system prompt, \textit{emotion} is a categorical label such as ``frustrated'' or ``satisfied,'' \textit{outcome} is the distilled policy reflection, and \textit{context} encodes optional structured metadata (e.g., product SKU or budget).  
The document is then embedded using BAAI/bge-m3 into a 768-dimensional vector. The collection supports two main operations: adding an Orb (which inserts the vector representation with metadata) and retrieving the top-$k$ most relevant Orbs based on similarity.

\paragraph{Reflection Pipeline.} After an episode finishes, the agent calls the LLM to generate a reflection based on the last $m$ Orbs. The LLM returns a concise paragraph, which is appended to the prompt of the next turn. During inference, the current user message is embedded and the most relevant reflections are retrieved and prepended to the context window.

\paragraph{Retrieval Pipeline.} 
Retrieval proceeds in three lightweight stages. Given the current user query $q$ and the running dialogue context $C$, we first prompt the frozen LLM to produce a memory query $q'$ that compresses $q$ and $C$ into a concise, retrieval-oriented question:
\begin{equation}
q' = \texttt{LLM}_{\text{rewrite}}(q \oplus C) 
\end{equation}
We then embed this query using the same BGE-M3 encoder employed at indexing time, yielding
\begin{equation}
\mathbf{e}_{q'} = \text{Embed}(q') \in \mathbb{R}^{768} 
\end{equation}
Finally, a maximum-inner-product search is performed over the ChromaDB collection of Orb embeddings to retrieve the top-$k$ most relevant Orbs:
\begin{equation}
\mathcal{R} = \operatorname*{top\text{-}k}_{i}\; \mathbf{e}_{q'} \cdot \mathbf{e}_{\mathcal{O}_i} 
\end{equation}
where $\mathcal{R}$ contains the IDs, text, and metadata of the selected Orbs, which are then concatenated and inserted into the agent prompt.

\begin{table}[t]
\centering
\caption{Success rate (\%) on \textsc{ECom-Bench}(Household appliances environment 53 tasks).}
\label{tab:model_success_rates_story}
\resizebox{\textwidth}{!}{%
\begin{tabular}{lcccccccccc}
\toprule
\textbf{Model} & \textbf{T1} & \textbf{T2} & \textbf{T3} & \textbf{T4} & \textbf{T5} & \textbf{T6} & \textbf{T7} & \textbf{T8} & \textbf{T9} & \textbf{T10} \\
\midrule
Doubao-Seed-1.6-Thinking & 45.18 & 69.81 & 75.47 & 84.91 & 84.91 & 84.91 & 84.91 & 86.79 & 88.68 & 88.68 \\
Doubao-Seed-1.6-Thinking-MemOrb & 32.08 & \textbf{62.26} & \textbf{69.81} & \textbf{83.02} & \textbf{88.68} & \textbf{90.57} & \textbf{90.57} & \textbf{92.45} & \textbf{94.34} & \textbf{94.34} \\
\midrule
Doubao-Seed-1.5 & 18.87 & 33.96 & 47.17 & 52.83 & 58.49 & 66.04 & 66.04 & 66.04 & 66.04 & 67.92 \\
Doubao-Seed-1.5-MemOrb & \textbf{32.08} & \textbf{60.38} & \textbf{77.36} & \textbf{81.13} & \textbf{88.68} & \textbf{88.68} & \textbf{88.68} & \textbf{92.45} & \textbf{94.34} & \textbf{94.34} \\
\midrule
Deepseek-V3 & 24.53 & 39.62 & 45.28 & 47.17 & 54.72 & 56.60 & 56.60 & 60.38 & 64.15 & 66.04 \\
Deepseek-V3-MemOrb & \textbf{28.30} & \textbf{52.83} & \textbf{60.38} & \textbf{67.92} & \textbf{73.58} & \textbf{73.58} & \textbf{75.47} & \textbf{75.47} & \textbf{75.47} & \textbf{75.47} \\
\bottomrule
\end{tabular}}
\end{table}

\begin{table}[t]
\centering
\caption{Success rate (\%) on \textsc{ECom-Bench}(Clothing items environment 77 tasks).}
\label{tab:model_success_rates_clothing}
\resizebox{\textwidth}{!}{%
\begin{tabular}{lcccccccccc}
\toprule
\textbf{Model} & \textbf{T1} & \textbf{T2} & \textbf{T3} & \textbf{T4} & \textbf{T5} & \textbf{T6} & \textbf{T7} & \textbf{T8} & \textbf{T9} & \textbf{T10} \\
\midrule
Doubao-Seed-1.6-Thinking & 10.39 & 20.78 & 23.38 & 24.68 & 28.57 & 29.87 & 31.17 & 32.47 & 36.36 & 37.66 \\
Doubao-Seed-1.6-Thinking-MemOrb & \textbf{12.99} & \textbf{22.08} & \textbf{28.57} & \textbf{31.17} & \textbf{31.17} & \textbf{33.77} & \textbf{35.06} & \textbf{35.06} & \textbf{38.96} & \textbf{38.96} \\
\midrule
Doubao-Seed-1.5 & 20.78 & 23.38 & 25.97 & 28.57 & 31.17 & 32.47 & 33.77 & 33.77 & 35.06 & 35.06 \\
Doubao-Seed-1.5-MemOrb & 20.78 & \textbf{29.87} & \textbf{29.87} & \textbf{32.47} & \textbf{35.06} & \textbf{35.06} & \textbf{37.66} & \textbf{37.66} & \textbf{37.66} & \textbf{37.66} \\
\midrule
Deepseek-V3 & 18.18 & 20.78 & 23.38 & 28.57 & 28.57 & 32.47 & 32.47 & 32.47 & 33.77 & 33.77 \\
Deepseek-V3-MemOrb & 18.18 & \textbf{25.97} & \textbf{28.57} & \textbf{29.87} & \textbf{31.17} & \textbf{33.77} & \textbf{36.36} & \textbf{36.36} & \textbf{36.36} & \textbf{36.36} \\
\bottomrule
\end{tabular}}
\end{table}
\section{Experiments}
\label{sec:experiments}

We evaluate \textbf{MemOrb} on \textsc{ECom-Bench}, a public simulator that covers 130 customer-service tasks spanning electronics (53 tasks) and clothing (77 tasks).  
\subsection{ECom-Bench Benchmark}
ECom-Bench~\citep{wang2025ecombenchllmagentresolve} is an e-commerce customer service simulation environment, consisting of a total of 130 tasks (including 53 household appliance tasks from the original benchmark and 77 newly introduced clothing tasks contributed in this work). These tasks are constructed based on hundreds of user data records and diverse user profiles. Each task is a multi-turn dialogue, where users raise questions covering aspects such as orders, logistics, and product knowledge. Success is binary: the agent must satisfy the customer request within 12 turns without hallucinating product or order information.

In this environment, an Agent equipped with encapsulated MCP tools must determine which tool to call based on the user’s query and input parameters. However, tool invocation can encounter several issues. One common issue is \textbf{incorrect tool selection}; for example, when the user provides a product ID or name without an order number, the model may mistakenly call the order query tool instead of the ``user order query tool,'' which can retrieve order numbers under the user’s ID. Another problem arises when the \textbf{correct tool is selected but erroneous input parameters are provided}, leading to failure in execution. Finally, even if both the tool selection and parameter passing are correct, there can still be a failure to \textbf{incorporate the returned results into the final response}, which undermines the effectiveness of the interaction.

Under such circumstances, the Agent is required to perform a reflection after each task, involving elements such as interaction trajectory, observation, outcome, and reflection. For subsequent similar tasks, the Agent should retrieve relevant reflections from a database based on the user’s query, and inject them into the system prompt to construct an enhanced prompt.

\subsection{Setup}
\label{subsec:setup}

\paragraph{Agent.}
We adopt the official \textsc{LangGraph} template agent shipped with \textsc{ECom-Bench}.  The LLM backbone is \textsc{Doubao-Seed-1.6-thinking}; parameters are frozen for the entire study.

\paragraph{Memory configurations.}  
The \textbf{No-Memory} configuration is the vanilla agent, with a 4k-token context. The \textbf{MemOrb} configuration is the same agent, but with \textbf{MemOrb} integrated. In this case, reflections are stored in SQLite, and vectors are managed in ChromaDB, with a retrieval parameter of $k = 5$ and an embedding dimension of 768.

\paragraph{Protocol.}
We run 10 independent \emph{trials}.  In trial-1 both agents start from scratch.  After each trial, \textbf{MemOrb} writes all new trajectories plus their reflections to the shared memory; the next trial begins immediately without resetting the memory bank.  No gradient updates occur at any point.

\subsection{Main Results}
\label{subsec:results}

On \textsc{ECom-Bench}, our MemOrb framework consistently improves multi-trial success rates across domains. 
In the \textbf{Household appliances environment (53 tasks)}, shown in Table~\ref{tab:model_success_rates_story}, Doubao-Seed-1.6-Thinking-MemOrb achieves a final success rate of \textbf{94.34\%}, notably higher than the baseline Doubao-Seed-1.6-Thinking (88.68\%). Similar improvements are observed for Doubao-Seed-1.5 and Deepseek-V3 when enhanced with MemOrb. 
In the \textbf{Clothing items environment (77 tasks)}, as reported in Table~\ref{tab:model_success_rates_clothing}, MemOrb also yields consistent gains, with DeepSeek-V3-MemOrb reaching \textbf{36.36\%} at T10 compared to 33.77\% for the baseline. Despite the relatively lower absolute performance in clothing tasks due to their higher complexity, MemOrb delivers stable relative improvements across models.

\subsection{Ablation Studies}
\label{subsec:ablation}
\begin{figure*}[t]
  \centering
  \includegraphics[width=\textwidth]{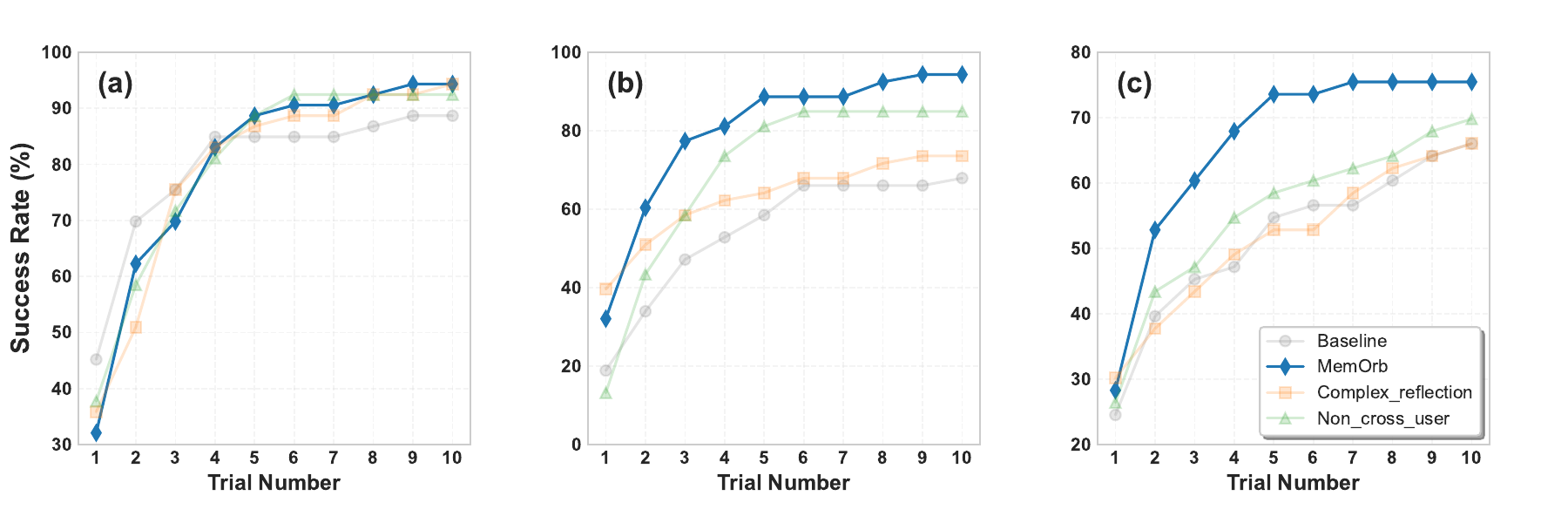}
  \caption{Ablation experiments on removing cross-user and adding complex structured reflection logic in 53 household appliance category tasks on ECom-Bench: (a) Doubao-Seed-1.6-Thinking. (b) Doubao-Seed-1.5. (c) DeepSeek-V3.}
  \label{fig:memory_ablation}
\end{figure*}

\paragraph{Complex memory ablation(Household Appliances Category).}
We conducted a comparison between complex structured reflection memory and Orb memory. As shown in Figure~\ref{fig:memory_ablation}, after introducing complex structured reflection memory, the effect does not show a significant improvement compared to the baseline. Moreover, it increases a large amount of token consumption and context occupation, which leads to higher costs, increased memory usage, and longer retrieval and response times. This is something that must be strongly avoided in large-scale multi-turn buyer-customer service conversations in the e-commerce customer service field.

\paragraph{Cross-user memory ablation(Household Appliances Category).}
We compared the impact of different models when using MemOrb versus removing the cross-user retrieval module (which is equivalent to changing the top-k parameter: when k is very small there is a higher chance of loading identical or highly similar memories — here we set k = 1, while in MemOrb k = 5; this choice balances retrieval efficiency and the relative diversity of Orbs), as shown in Figure \ref{fig:memory_ablation}.

At the model level, using the deep-thinking model Doubao-Seed-1.6-Thinking has little effect on the cross-user memory ablation. However, Doubao-Seed-1.5 and DeepSeek-V3 exhibit degraded performance without cross-user memory compared to the success rate (SR) achieved with MemOrb’s cross-user memory.

Although removing cross-user memory still yields a success-rate curve that rises above the baseline, the agent system tends to get trapped in local optima: after multiple failures it cannot escape difficult situations, and after repeated successes it lacks the ability to transfer or generalize. This highlights the importance of cross-user memory retrieval in MemOrb.

In large-scale e-commerce customer-service settings, using fast, non-deep-thinking models is beneficial for improving efficiency and reducing costs, which further validates the feasibility of MemOrb.

\paragraph{Change the indicator for evaluating task completion rate.}
To verify the multi-trial success rate of MemOrb on the 53 household appliance categories in ECom-Bench, we adopt the evaluation metric $Pass^k$ introduced in~\citep{yao2024taubenchbenchmarktoolagentuserinteraction}, which is formally defined as
\begin{equation}
\text{pass}^k = \mathbb{E}_{\text{task}} \left[ \frac{\binom{c}{k}}{\binom{n}{k}} \right]
\end{equation}

\begin{figure*}[t]
  \centering
  \includegraphics[width=\textwidth]{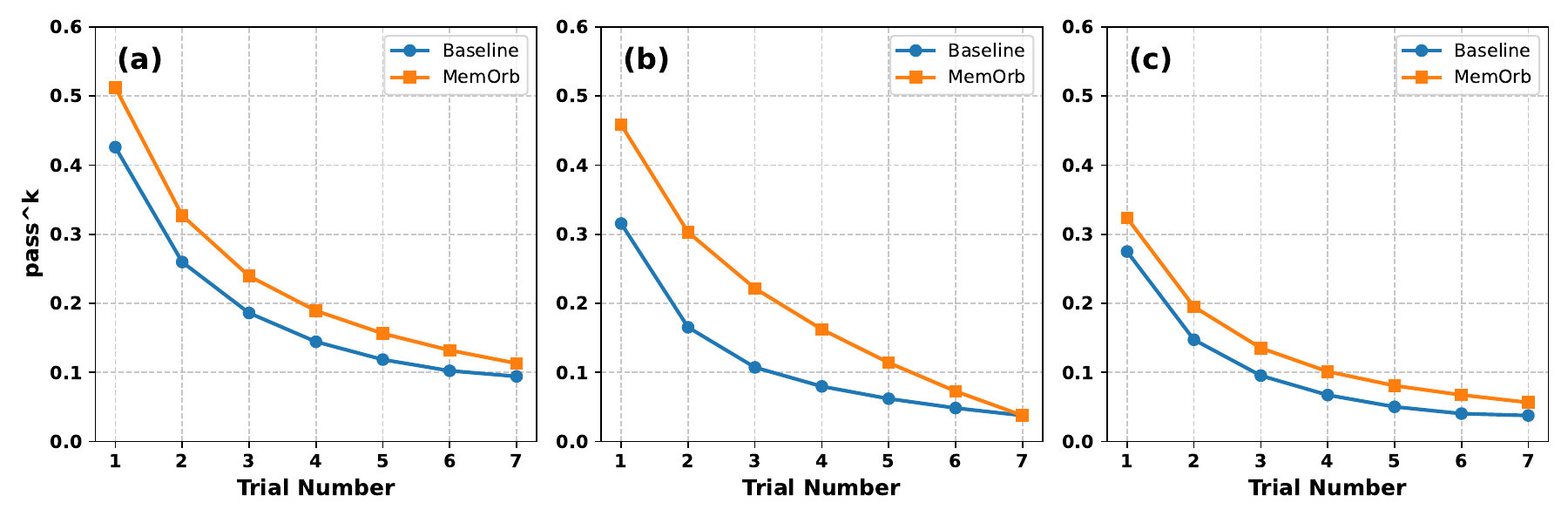}
  \caption{$Pass^k$ in 53 household appliance category tasks on ECom-Bench: (a) Doubao-Seed-1.6-Thinking. (b) Doubao-Seed-1.5. (c) DeepSeek-V3.}
  \label{fig:pass^k}
\end{figure*}

This metric represents the probability that all $k$ independent and identically distributed task trials succeed. It emphasizes consistency, requiring consecutive successes across multiple trials. Such a setting is particularly relevant for scenarios like batch processing or e-commerce customer service, where strict evaluation of model stability and reliability is necessary. Moreover, we modified the calculation logic by changing the original metric, which by default skipped correctly completed tasks and re-executed the same task, as illustrated in Figure~\ref{fig:pass^k}. It can be seen that compared with the baseline model, the use of MemOrb significantly improves its success rate in multiple consecutive tasks.

\subsection{Qualitative Analysis}
\label{subsec:qualitative}

Manual inspection of 50 conversations shows that \textbf{MemOrb} reduces repetitive clarifications and correctly re-uses prior refund-policy explanations, while the baseline repeats the same retrieval of knowledge-base articles across trials.

\subsection{Reproducibility}
\label{subsec:repro}
We release the exact SQLite schema, ChromaDB snapshots, and evaluation scripts at GitHub.  All numbers can be reproduced with a single \texttt{python run\_eval.py} command.
\section{Discussions}
\label{sec:discussion}

\paragraph{Limitations.}
\textbf{MemOrb} currently assumes that the LLM used for rewriting and reflection remains frozen.  While this guarantees zero-shot deployment, it also caps the quality of distilled reflections to the base model’s capability.  Additionally, our SQLite + ChromaDB stack is single-node; scaling to millions of concurrent sessions will require sharding and replication strategies that we leave for future work.

In the experimental section, we only conducted tests on the ECom-Bench benchmark in the e-commerce field and did not cover some other common benchmarks. Lack of multimodality: Orbs is limited to text and ignores e-commerce visual elements such as receipts and product information, which limits its authenticity.

\paragraph{Broader Impact.}
By making continual learning accessible without gradient updates, \textbf{MemOrb} lowers the barrier for small and medium businesses to adopt adaptive customer-service agents. The proposed cross-user reflection effectively avoids Agents from falling into local optima in the same or highly similar tasks, and improves retrieval efficiency through rewriting.

\paragraph{Future Directions.}
(1) \emph{Multi-modal memories}: extend Orbs to include receipts, screenshots, and voice snippets.  
(2) \emph{Privacy}: investigate federated or on-device storage to comply with GDPR/CCPA.  
(3) \emph{Cross-domain transfer}: evaluate MemOrb on healthcare and finance tasks to test generalisation beyond e-commerce.
\section{Conclusion}
\label{sec:conclusion}
In this work, we introduced \textsc{MemOrb}, a lightweight and schema-free verbal reinforcement memory layer that enables frozen LLM agents to achieve continual improvement without gradient updates. By distilling task trajectories into compact strategy reflections and storing them in a hybrid SQLite + ChromaDB architecture, MemOrb substantially enhances consistency and success rates across multi-turn customer-service tasks on \textsc{ECom-Bench}, achieving up to a 71 percentage-point improvement over strong baselines. 

Beyond demonstrating effectiveness in the e-commerce domain, MemOrb provides a general mechanism for cross-user knowledge transfer and stable long-term adaptation. Looking forward, we aim to extend MemOrb to multimodal settings (e.g., integrating receipts, screenshots, and voice records), investigate privacy-preserving storage via federated or on-device deployments, and explore its applicability in domains such as healthcare and finance. We release all code, data, and evaluation pipelines to encourage future research on self-evolving language agents.

\bibliography{main}
\bibliographystyle{iclr2026_conference}

\appendix
\section{EXPERIMENT DETAILS}
\subsection{Prompt}

\begin{tcolorbox}[
    colback=blue!5!white,
    colframe=blue!60!black,
    title=\textbf{Raw System Prompt},
    fonttitle=\bfseries,
    breakable,
    enhanced,
    attach boxed title to top left={xshift=1.5em, yshift=-2mm},
    boxed title style={colback=blue!60!black}
]
\textbf{Basic Information} \\
You are now an e-commerce customer service representative. The platform you belong to is \textbf{\{platform\}}, the shop\_id of your store is \textbf{\{shop\_id\}}, and the user\_id of the customer you serve is \textbf{\{user\_id\}}.
\end{tcolorbox}

\vspace{0.5cm}

\begin{tcolorbox}[
    colback=blue!5!white,
    colframe=blue!60!black,
    title=\textbf{System Prompt With Reflection Template},
    fonttitle=\bfseries,
    breakable,
    enhanced,
    attach boxed title to top left={xshift=1.5em, yshift=-2mm},
    boxed title style={colback=blue!60!black}
]
\textbf{Basic Information} \\
You are now an e-commerce customer service representative. The platform you belong to is \textbf{\{platform\}}, the shop\_id of your store is \textbf{\{shop\_id\}}, and the user\_id of the customer you serve is \textbf{\{user\_id\}}. You will be given a past task execution history, in which you were placed in an e-commerce customer service environment and given a task to complete. 

\vspace{0.3cm}

\textbf{Task Scenario} \\
\textcolor{blue}{\{scenario\_desc\}}

\vspace{0.3cm}

\textbf{Complete Execution History} \\
\textcolor{blue}{\{''.join(self.\_format\_messages())\}}

\vspace{0.3cm}

\textbf{Performance Evaluation Result} \\
- \textbf{Action tool score:} \textcolor{red}{\{action\_reward\}}/1.0 \\
- \textbf{Search tool score:} \textcolor{red}{\{search\_reward\}}/1.0 \\
- \textbf{Output score:} \textcolor{red}{\{output\_reward\}}/1.0 \\
- \textbf{Overall result:} \textcolor{red}{\texttt{"Success" if reward > 0.5 else "Failure"}} 

\vspace{0.3cm}

\textbf{Reflection on Historical Failures (if any)} \\
\textcolor{blue}{\{memory\_context if memory\_context else "No historical reflection"\}}

\vspace{0.3cm}

\textbf{Reflection Guidance} \\
Instead of summarizing the environment, delve deeply into the strategies and paths you've taken to attempt to complete the task. Develop a concise and innovative action plan, focusing on learning from failure experiences: \\
- Carefully analyze the error patterns in historical failure memories \\
- Identify the key factors and decision points that lead to failure \\
- Develop strategies to avoid repeating mistakes based on failure experiences

\vspace{0.3cm}

\textbf{Reflection Requirements (Focusing on Learning from Failure Experiences)} \\
1. \textbf{Conduct in-depth analysis of key decision points and failure causes during the task execution process} \\
2. \textbf{Identify specific issues in tool selection, parameter usage, and sequence arrangement} \\
3. \textbf{Develop specific improvement plans, focusing on failure experiences:} \\
    - \textcolor{blue}{Analyze common error patterns in historical failed memories} \\
    - \textcolor{blue}{Identify the key factors leading to failure} \\
    - \textcolor{blue}{Develop strategies to avoid repeating the same mistakes} \\
4. Focus on the identification and avoidance of failure modes. \\
5. Reflection should begin with \textbf{"I \{ 'succeeded' if reward > 0.5 else 'failed' \} in this mission"} \\
6. At the end of the reflection, clearly provide a summary of \textbf{"New Plan:"}, which must be based on historical failure experiences and focus on avoiding known error patterns. Please provide a detailed and actionable reflection summary, focusing on learning from failure experiences and making improvements.

\end{tcolorbox}

\subsection{Dialogue case}
\begin{tcolorbox}[
    colback=blue!5!white,
    colframe=blue!60!black,
    title=Example of the first failed conversation,
    fonttitle=\bfseries,
    breakable,
    enhanced,
    attach boxed title to top left={xshift=1.5em, yshift=-2mm},
    boxed title style={colback=blue!60!black}
]
\textbf{User}: I recently bought a product from you, and the link is this:\\ \textcolor{blue}{https://item.jd.com/100042754736.html?sdx=ehi-lLxFuJiE6JnIYIpei8AitzeRRHsgmjYZ4u\\kJEdyMdZnQK5xZ53jtoU8\&sdx=ehi-lLxFuJiE6JnIYIpei8AitzeRRHsgmjYZ4ukJEdyMdZ\\nQK5xY7njhp04}. The installation is extremely troublesome. Could you please improve it? 

\textbf{Assistant}: Dear customer, thank you very much for your feedback~ Regarding the Midea F6033-JE6(HE) electric water heater you purchased, the current installation process requires first confirming that the installation environment meets the requirements for power supply, water pressure, etc., and then proceeding with fixing, pipe connection, and debugging. The basic installation fee is 100 yuan, and it also includes a standard accessory package and the first debugging service. Could you please tell us which specific step in the installation process you find troublesome? Do you need additional assistance or have suggestions for optimizing the process? You can share the details with us, and we will promptly provide feedback to the relevant department. 

\textbf{User}: I find the entire installation process quite troublesome, especially the steps of fixing and connecting. Do you have a detailed installation tutorial? I'd like to see if I can solve the problem myself. 

\textbf{Assistant}: Dear customer, we fully understand the troubles you've encountered during the installation process~ Currently, information on the detailed installation tutorial for this product is missing, so we're temporarily unable to provide specific tutorial content. However, you can contact our professional installation team, and they will provide you with one-on-one guidance~ Would you like us to arrange for an installation technician to assist you again or answer your questions? 

\textbf{User}: Okay, could you please help me contact the installation technician? Also, I'm going on a trip soon. Could you send me the logistics information for order \textbf{313021098954}? Please buzz the processing. 

\textbf{Assistant}: Your Highness, here is the logistics information for your order \textbf{313021098954}: 
\textbf{Logistics Tracking Number: JD1234567890} 
\textbf{Logistics Company: JD Express} 
\textbf{Estimated Delivery Time: 2025-04-07 14:00:00} 
\textbf{Current Status: Created (08:30 on April 1, 2025, Gulou District, Nanjing)} 
We have processed this order with priority and will arrange for shipment as soon as possible~ 

Regarding contacting the installation technician, we need you to provide the following information to make an appointment for you: 
\begin{itemize} 
    \item Service type (such as "installation guidance", "on-site inspection") 
    \item Convenient service time (Monday to Sunday) 
    \item Your contact phone number 
\end{itemize} 
We will immediately register and make an appointment for you once you provide the information. 

\textbf{User}: Okay, the service type is installation guidance. The convenient service time is this weekend afternoon. Please check my contact number for me; I can't remember it. Additionally, I would like to check the status of order \textbf{313271663680}. 

\textbf{Assistant}: Your Grace, the status of your order \textbf{313271663680} is Paid. The total amount of the order is 3294 yuan, including the following products: 
\textbf{Product ID 100112573625} (unit price 1416 yuan, quantity 1) 
\textbf{Product ID 100040350131} (unit price 939 yuan, quantity 2) 
The delivery address is No. 231, North Zhongshan Road, Raoping County, Jieyang City, Guangdong Province, and the contact phone number is \textbf{15540534614}~ 

Regarding the installation guidance service, we have found the contact number \textbf{15540534614} for you. The "weekend afternoon" you mentioned needs to be specifically confirmed as Saturday afternoon or Sunday afternoon. Once Your Grace confirms, we will immediately make an appointment with the installation master for you. 

\textbf{User}: Okay, the specific time is Saturday afternoon. If order \textbf{313271663680} has not been shipped yet, I would like to cancel this order. 

\textbf{Assistant}: Dear customer, the current status of your order \textbf{313271663680} is Paid, which meets the cancellation conditions. The order details are as follows: 
\textbf{Order amount: 3294 yuan} 
\textbf{Product information:} 
\begin{itemize} 
    \item Product ID 100112573625 (unit price 1416 yuan, quantity 1) 
    \item Product ID 100040350131 (unit price 939 yuan, quantity 2) 
\end{itemize} 
\textbf{Payment method: WeChat} 
Please reply with \textbf{"Confirm"} to proceed with canceling the order. After cancellation, the payment will be refunded via the original payment path. 

\textbf{User}: Confirm cancellation of order \textbf{313271663680}. Additionally, I would like to ask if there is any cashback record for order \textbf{314231443863}? 

\textbf{Assistant}: Your order \textbf{313271663680} has been successfully canceled, my lady. The payment will be refunded through the original payment method (WeChat), and it is expected to arrive within 1-3 working days. 

In addition, it is found that there is currently no record of cashback for sharing the order review for order \textbf{314231443863} (status: no order review shared). If you have already shared the order review, you can provide the proof of sharing (such as a screenshot link), and I will verify it for you and register the cashback process. 

\textbf{User}: Okay, I have indeed posted a product review. Here is the link to my review image: \\\textcolor{blue}{https://dd-static.jd.com/ddimgp/jfs/t20260528/280781/10/25581/172166/6808a980F4fea48\\67/cf783a9a7acc8c2d.jpg}. Please help me verify it and register the cashback information. 

\textbf{Assistant}: Your Highness, I'm very sorry~ The review content from the user is not included in the screenshot of the order sharing you provided, so it can't be verified as a valid proof of order sharing for the time being. Could you please provide a link to the complete screenshot that includes the review? I will help you recheck and register for the cashback process. 

\textbf{User}: Okay, I will re-provide the complete screenshot containing the comments. Thank you for your help.

\textbf{trial: 1}

\textbf{detail\_reward:}\\
- \textbf{reward: \textcolor{red}0}\\
- \textbf{action: \textcolor{red}0}\\
- \textbf{search: \textcolor{red}1}\\
- \textbf{output: \textcolor{red}1}
\end{tcolorbox}

\begin{tcolorbox}[
    colback=blue!5!white,
    colframe=blue!60!black,
    title=Reflection Example,
    fonttitle=\bfseries,
    breakable,
    enhanced,
    attach boxed title to top left={xshift=1.5em, yshift=-2mm},
    boxed title style={colback=blue!60!black}
]
\textbf{Task ID: 0}

\textbf{Memory:}

I failed this task. When handling the user's request regarding the cancellation of order \textbf{313271663680} and cashback for sharing order details, there were issues with tool invocation and process handling.  

\textbf{Example:}  

\texttt{ "name": "get\_image\_info\_tool", "parameters": \{ "history\_messages": "Yes, I have indeed posted a product review. Here is the link to my review image: \textcolor{blue}{https://dd-static.jd.com/ddimgp/jfs/t20260528/280781/10/25581/\\172166/6808a980F4fea4867/cf783a9a7acc8c2d.jpg}. Please help me verify it and register the cashback information.", "summarized\_query": "Verify the validity of the product review screenshot to register cashback", "needed\_query": "Does the screenshot contain the user's comment?" \} }

\textbf{Improvement Plan:}

\begin{itemize}
    \item \textbf{Order Cancellation Process:} Directly call \\\texttt{manage\_order\_tool(action=cancel, order\_id=313271663680)} without redundant queries for user information.
    \item \textbf{Product Review Verification Process:} Call \texttt{get\_image\_info\_tool} to extract the comment content from the image and verify the validity of the product review.
    \item \textbf{Refund Judgment Process:} Check the payment method after canceling an order, and only call \texttt{manage\_ecard} for a refund when the payment method is \textbf{"JD E-card"}.
\end{itemize}

\textbf{Refund Process Judgment After Order Cancellation:}
\begin{itemize}
    \item After canceling an order, the payment method must be checked:
    \begin{itemize}
        \item If it is \textbf{"JD E-card"}, \\immediately call \texttt{manage\_ecard(action=refund, ...)}.
        \item If it is other payment methods (such as WeChat), no additional action is required. This judgment logic must be clearly reflected in the tool call.
    \end{itemize}
\end{itemize}

\textbf{Summary of the New Plan:}

\begin{itemize}
    \item \textbf{Order cancellation process:} Directly call \\\texttt{manage\_order\_tool(action=cancel, order\_id=313271663680)}.
    \item \textbf{Product review verification process:} Call \texttt{get\_image\_info\_tool} to extract and verify the review.
    \item \textbf{Refund judgment process:} Check the payment method after canceling the order, and only call \texttt{manage\_ecard} if the payment method is \textbf{"JD E-card"}.
\end{itemize}

Through the above improvements, tool call errors and process redundancy can be avoided, ensuring strict adherence to operating rules and improving the accuracy of task execution.

\textbf{Status:} Success  

\textbf{Last Update:} \texttt{2025-09-05 15:22:44.418306}
\end{tcolorbox}

\begin{tcolorbox}[
    colback=blue!5!white,
    colframe=blue!60!black,
    title={After reflecting on the first failure, the task was successfully handled the second time},
    fonttitle=\bfseries,
    breakable,
    enhanced,
    attach boxed title to top left={xshift=1.5em, yshift=-2mm},
    boxed title style={colback=blue!60!black}
]

\textbf{Assistant}: Hi, is there anything I can help you with?

\textbf{User}: I recently bought a product from you, and installing it is particularly troublesome. It's really frustrating. Here's the link to the product:\\ \textcolor{blue}{https://item.jd.com/100042754736.html?sdx=ehi-lLxFuJiE6JnIYIpei8AitzeRRHsgmjYZ4u\\kJEdyMdZnQK5xZ53jtoU8\&sdx=ehi-lLxFuJiE6JnIYIpei8AitzeRRHsgmjYZ4ukJEdyMdZ\\nQK5xY7njhp04.}

\textbf{Assistant}: Your Grace, I'm very sorry for the unpleasant installation experience and making you angry~ I'll immediately check the installation instructions for this product for you: \textbf{Before installation}, please ensure that the \textbf{power supply is 220V}, the \textbf{socket is well grounded}, and the \textbf{water pressure} is between \textbf{0.05MPa} and \textbf{0.6MPa}. The \textbf{installation location} should be away from flammable materials and well-ventilated. The \textbf{basic installation fee is 100 yuan}, which includes fixing the water heater, connecting the water pipes, and power-on debugging. For water pipes exceeding the standard length, an additional \textbf{20 yuan per meter} will be charged. Fees for modifying water and electrical circuits will be calculated separately. Would you like me to provide you with more detailed installation guidance, or help you contact a professional installer for assistance?

\textbf{User}: Okay, please send me a detailed installation tutorial first. Let me see if I can handle it myself.

\textbf{Assistant}: Your Highness, I completely understand the troubles you encountered during installation. Here is a \textbf{detailed installation tutorial} for this Midea electric water heater for you~

\textbf{Preparation Before Installation}

- \textbf{Environmental Requirements}:
  - \textbf{Power supply}: \textbf{220V} dedicated socket (must be grounded, avoid sharing with other high-power electrical appliances).
  - \textbf{Water pressure}: \textbf{0.05MPa-0.6MPa} (you can contact the property management to check if the water pressure meets the standard).
  - \textbf{Location}: Keep away from flammable materials (such as gas pipes, paint, etc.), avoid direct sunlight, and reserve maintenance space (it is recommended to keep a distance of more than \textbf{10cm} from the wall).
  
- \textbf{Tools and Accessories}:
  - Free accessory package: \textbf{Expansion bolts 2}, \textbf{sealing rings 2}, \textbf{water inlet and outlet pipes 1} (within \textbf{1.5 meters} in length).
  - Self-prepared tools: \textbf{Electric drill}, \textbf{Phillips screwdriver}, \textbf{wrench}, \textbf{level} (to ensure the machine is installed vertically).

\textbf{Step-by-Step Installation Process}

1. \textbf{Mark Installation Holes}:  
   Use a level to draw a horizontal line for the water heater's hanging plate on the wall, and mark two drilling points according to the distance between the hanging plate holes (\textbf{about 40cm}).  
   After drilling with an electric drill, insert expansion tubes and fix the hanging plate with bolts (ensure the hanging plate is firm and can bear a weight of more than \textbf{50kg}).

2. \textbf{Fix the Water Heater}:  
   Two people work together to hang the water heater on the hanging plate, gently push the machine to check if it shakes (if loose, the hanging plate needs to be readjusted).

3. \textbf{Connect Water Inlet and Outlet Pipes}:  
   Distinguish between \textbf{cold} and \textbf{hot water ports} (marked \textbf{"cold"} and \textbf{"hot"} on the machine). Connect the water inlet pipe (blue interface) to the cold water valve, and the water outlet pipe (red interface) to the hot water pipe.  
   Tighten the interfaces with a wrench and put on the sealing rings to prevent water leakage (it is recommended to wrap \textbf{raw tape} to enhance sealing).

4. \textbf{Power-on Debugging}:  
   Turn on the cold water valve and wait for the water heater to be filled with water (about \textbf{5 minutes}; you can confirm that the air is exhausted by draining water from the water outlet).  
   Plug in and turn on the machine, press the "power button" to start it. The water temperature is displayed by default, and you can adjust the temperature through the "\textbf{+/-}" buttons (it is recommended to set it to \textbf{50-60 degree Celsius} for energy saving and safety).

\textbf{Precautions}:

- Do not modify the circuit/water circuit by yourself: If you need to extend the water pipe or adjust the socket position, please contact a \textbf{professional electrician/plumber} (additional fees apply, you can consult customer service for quotes).
- \textbf{Exhaust before first use}: Make sure the water tank is full of water before starting the machine; otherwise, the heating tube may be damaged.
- \textbf{Free service}: For the first installation, you can make an appointment for a master to visit (basic installation fee is \textbf{100 yuan} including labor; additional accessories require payment). Appointment phone: \textbf{400-889-9315} (Midea official after-sales service).
- If you encounter specific problems during installation (such as water leakage at the interface, startup errors, etc.), you can tell me at any time, and I will help you further troubleshoot~ Do you need me to help you make an appointment for an on-site installation master?

\textbf{User}: Thank you for the detailed tutorial. I'll try it myself first. By the way, I have another order with the number \textbf{313021098954}. I'm going to travel soon. Could you please check the logistics information for me and handle it with urgency?

\textbf{Assistant}: Your Highness, we have checked the logistics information for order \textbf{313021098954} for you:  
- \textbf{Logistics Tracking Number}: \textbf{JD1234567890}  
- \textbf{Logistics Company}: JD Express  
- \textbf{Current Status}: Created (\textbf{08:30 on April 1, 2025, Gulou District, Nanjing})  
- \textbf{Estimated Delivery}: \textbf{14:00 on April 7, 2025}  
We have also submitted an \textbf{urgent processing} application for you, and the warehouse will prioritize the shipment~ You can check real-time logistics updates through \textbf{"My Orders"} on the JD APP. Feel free to let me know if you have any other questions~

\textbf{User}: Okay, thank you for the urgent processing. Could you also help me check the status of another order, \textbf{313271663680}?

\textbf{Assistant}: Your Highness, we have checked the status of order \textbf{313271663680} for you as follows:  
- \textbf{Order Status}: Paid  
- \textbf{Order Amount}: \textbf{3294 yuan}  
- \textbf{Product Information}:
  - Product ID: \textbf{100112573625}, unit price \textbf{1416 yuan}, quantity \textbf{1 piece}
  - Product ID: \textbf{100040350131}, unit price \textbf{939 yuan}, quantity \textbf{2 pieces}  
- \textbf{Shipping Address}: \textbf{No. 231, Zhongshan North Road, Raoping County, Jieyang City, Guangdong Province} (Recipient: \textbf{Song Jingming}, Phone: \textbf{155****4614})  
- \textbf{Payment Method}: WeChat  
The current order status is \textbf{normal}. The logistics information will be updated after the order is shipped. You can check the order status at any time~

\textbf{User}: Has this order not been shipped yet? If it hasn't been shipped, I want to cancel this order.

\textbf{Assistant}: My lady, your order \textbf{313271663680} has been successfully \textbf{canceled}, and the current status of the order is \textbf{"Cancelled"}. The order amount of \textbf{3294 yuan} will be refunded via the original payment method (WeChat). It is expected to arrive within \textbf{1-3 working days}. Please keep an eye on your account information~

\textbf{User}: Okay, thank you for your help. There's another order with the number \textbf{314231443863}, and I want to know if there's a cashback record for this order?

\textbf{Assistant}: Dear customer, after checking, the current status of the review and cashback for order \textbf{314231443863} is \textbf{"No review submitted"}. There is no review record for this order in the system yet. If you have already completed the review, you can provide the review certificate (such as a screenshot link), and I will help you verify it and register the cashback process~

\textbf{User}: Okay, I have indeed shared my order. Here is the screenshot link of my order sharing:\\ \textcolor{blue}{https://dd-static.jd.com/ddimgp/jfs/t20260528/280781/10/25581/172166/6808a980F4fea486\\7/cf783a9a7acc8c2d.jpg} Please help me verify and register the cashback information.

\textbf{Assistant}: Your Highness, the screenshot of your order review you provided has been verified as valid and contains user comment content. We have successfully registered the cashback for the order review of order \textbf{314231443863} for you. The cashback process has been triggered, and the cashback amount is expected to be returned to your payment account via the original route within \textbf{1-3 working days}. Please keep an eye out for it~

\textbf{User}: Okay, thank you for your help. I will pay attention to receiving the cashback.

\textbf{trial: 2}

\textbf{detail\_reward:}\\
- \textbf{reward: \textcolor{red}1}\\
- \textbf{action: \textcolor{red}1}\\
- \textbf{search: \textcolor{red}1}\\
- \textbf{output: \textcolor{red}1}

\end{tcolorbox}

\section{77 clothing category data codes constructed based on ECom-Bench}

\subsection{Master data generation Python script example}

\begin{lstlisting}[style=custompython]
class UserServiceInvoiceExtract(BaseModel):
    think: str = Field(..., description=(
        "Please explain in detail how you think, analyze, and reason based on the chat content, "
        "user needs, and background information to decide how to generate user, service appointment, "
        "and invoice information."
    ))
    user_info: UserBase = Field(..., description="User information")
    service_info: ServiceAppointment = Field(..., description="Service appointment information")
    invoice_info: InvoiceBase = Field(..., description="Invoice information")


class OrderExtract(BaseModel):
    order_counts: str = Field(..., description=(
        "Based on the provided information, analyze how many orders need to be generated and give reasons; "
        "when there are multiple order numbers, prioritize generating the corresponding multiple order information."
    ))
    order_think: str = Field(..., description=(
        "Please explain in detail how you think, analyze, and reason based on the chat content, user needs, "
        "and background information to decide how to generate each order. This should reflect your understanding "
        "and judgment of various factors such as products, quantity, user identity, shipping address, logistics "
        "method, order time, etc., and explain your generation basis, reasoning logic, normative constraints, "
        "and final generation strategy. Your thought process should help others understand why you designed the order "
        "this way, rather than simply listing the steps."
    ))
    order_info: List[OrderBase] = Field(..., description="Information for each order")


class ProductExtract(BaseModel):
    product_counts: str = Field(..., description="Based on the provided information, analyze how many products need to be generated and give reasons.")
    product_think: str = Field(..., description=(
        "Please explain in detail how you think, analyze, and reason based on the chat content, user needs, "
        "and background information to decide how to generate each product. This should reflect your understanding "
        "and judgment of various factors such as product type, quantity, attributes, brand, price, gifts, materials, "
        "etc., and explain your generation basis, reasoning logic, normative constraints, and final generation strategy. "
        "Your thought process should help others understand why you designed the product content this way, rather than "
        "simply listing the steps."
    ))
    product_info: ClothingProductBase = Field(..., description="Product information")


class LogisticsExtract(BaseModel):
    logistics_counts: str = Field(..., description="Based on the provided information, analyze how many logistics entries need to be generated and give reasons.")
    logistics_think: str = Field(..., description=(
        "Please explain in detail how you think, analyze, and reason when generating logistics information based on "
        "the chat content, user needs, and background information. This should reflect your understanding of and judgment "
        "on various elements such as orders, products, user shipping address, logistics company, delivery method, time points, "
        "logistics status, etc., and explain how you determine the basis, reasoning logic, rule constraints, and final design "
        "for each logistics entry. Your thought process should help others understand why you constructed the logistics "
        "information this way, rather than simply listing the steps."
    ))
    logistics_info: ClothingLogisticsBase = Field(..., description="Logistics information")


class ExtractLLM(LLM):
    def __init__(self, model_name, target, temperature=1) -> None:
        super().__init__(model_name=model_name, verbose=True, temperature=temperature)
        self._initiate_llm()
        self.parser = PydanticOutputParser(pydantic_object=target)

    @override
    def call(self, background_info: str) -> Dict:
        prompt_template = PromptTemplate(
            template="""
Task Requirements:
1. Assume you are responsible for synthesizing the required information in the e-commerce domain.
2. You need to refer to the available information and synthesize data accordingly.
3. Order and logistics information can only be referenced as many fields are missing; please be creative and enrich the details as much as possible.
4. Exchangeable products must come from the candidate product list, and only the most relevant products should be selected.
5. No product outside the list should be added as an exchangeable option.
6. Logistics information must match the orders: unshipped orders should have no logistics information; orders that are shipped but not signed should not show "signed", and orders marked as signed should end with "signed".

# Background Information:
{background_info}

# Output Format:
{format_instructions}""",
            input_variables=["background_info"],
            partial_variables={"format_instructions": self.parser.get_format_instructions()},
        )
        prompt = prompt_template.format_prompt(background_info=background_info)
        response = self.llm.invoke(prompt).content
        parsed_response = self.parser.parse(response)
        return parsed_response.model_dump()

    @override
    def load_system_prompt(self, system_prompt):
        return super().load_system_prompt(system_prompt)
\end{lstlisting}

\end{document}